# New Capability to Look Up an ASL Sign from a Video Example

Carol Neidle,[1] Augustine Opoku,[1] Carey Ballard,[1]
Yang Zhou,[2] Xiaoxiao He,[2] Gregory Dimitriadis,[2] Dimitris Metaxas[2]

[1]Boston University, [2]Rutgers University

**Abstract**. Looking up an unknown sign in an ASL dictionary can be difficult. Most ASL dictionaries are organized based on English glosses, despite the fact that (1) there is no convention for assigning English-based glosses to ASL signs; and (2) there is no 1-1 correspondence between ASL signs and English words. Furthermore, what if the user does not know either the meaning of the target sign or its possible English translation(s)? Some ASL dictionaries enable searching through specification of articulatory properties, such as handshapes, locations, movement properties, etc. However, this is a cumbersome process and does not always result in successful lookup. Here we describe a new system, publicly shared on the Web, to enable lookup of a video of an ASL sign (e.g., a webcam recording or a clip from a continuous signing video). The user submits a video for analysis and is presented with the five most likely sign matches, in decreasing order of likelihood, so that the user can confirm the selection and then be taken to our ASLLRP Sign Bank entry for that sign. Furthermore, this video lookup is also integrated into our newest version of SignStream® software to facilitate linguistic annotation of ASL video data, enabling the user to directly lookup a sign in the video being annotated, and, upon confirmation of the match, to directly enter into the annotation the gloss and features of that sign, greatly increasing the efficiency and consistency of linguistic annotations of ASL video data.

**Introduction.** There are many ASL dictionaries online (some freely accessible and some accessible by paid subscription) and in print. However, looking up a sign can be difficult. Most ASL dictionaries are organized based on English glosses, despite the fact that (1) there is no convention for assigning English-based glosses to ASL signs; and (2) there is no 1-1 correspondence between ASL signs and English words. Furthermore, what if the user does not know either the meaning of the target sign or its possible English translation(s)? Some ASL dictionaries enable searching through specification of articulatory properties, such as handshapes, locations, movement properties, etc. For example, as illustrated in Figure 1 for BOOK, https://www.handspeak.com/word/asl-eng/ offers search based on parameters.

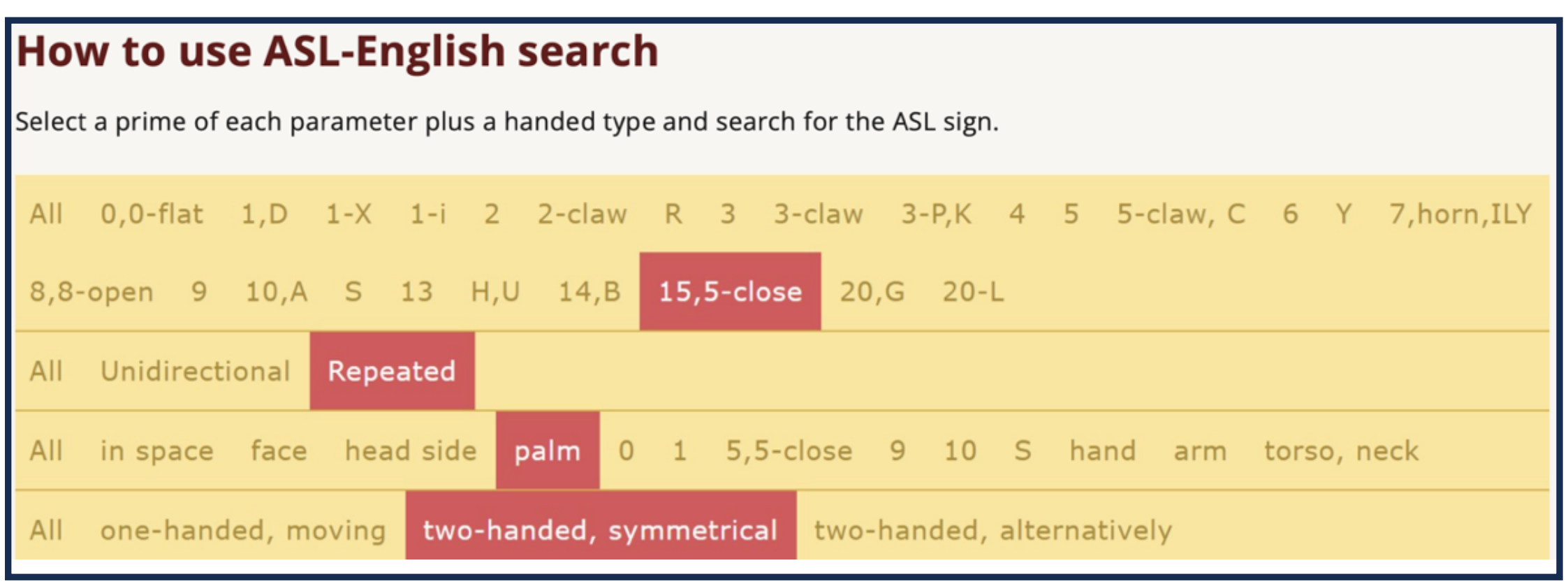

*Figure 1. Search Interface from handspeak.com*

However, that comes with this caveat (as of 7/14/24):

> "This is a pilot / ongoing development. It is subject to revisions or adjustments. Not all words have been updated for this reverse dictionary."

However, even if this were complete, it is cumbersome to search this way.

It would certainly be desirable to be able to look up an unknown ASL sign based on an example video—either a webcam video of an individual target sign, or an unknown sign segmented from a sequence of signs (i.e., a sentence) in a video. This is now possible !!

**New Lookup Tool.** Using an AI approach described by Zhou, et al. (2024), we have trained on about 98,000 consistently labelled video examples (Neidle, et al. 2022a) to enable recognition of about 2,360 signs, as well as similar variants thereof. We achieve overall sign recognition accuracy of 80.8% Top-1 and 95.2% Top-5 for citation-form signs, and 80.4% Top-1 and 93.0% Top-5 for signs pre-segmented from continuous signing. On our website, the user can upload a video and submit it for sign recognition. The user will then be presented with their source video along with videos of the top 5 closest sign matches, in decreasing order of likelihood, as well as variants that may be available for any of those signs. The user can play any or all of those videos, and then confirm a final selection, if the correct sign is included in the set of 5 that are displayed (which is likely to occur most of the time, even if the recognition accuracy for non-native sign productions may be a bit lower than the results just reported). We delete any uploaded videos immediately after processing, to maintain privacy, although we keep statistics on the frequency with which users confirm that their target video is listed as the 1st, 2nd, 3rd, 4th, or 5th choice, or none of the choices, so that we can, over time, assess the success rate for the system.[1]

The user interface for submitting a video for lookup is shown in Figure 2.

---

[1] A user study (Xu*, et al.* 2022) conducted on *Gloss-Finder,* a similar kind of prototype system trained on the WLASL dataset of citation-form signs (Li*, et al.* 2020), showed that 10 hearing learners with different levels of ASL experience preferred using video-based lookup and found it more effective as compared with list-based or parameter-based methods for searching for an unknown sign. It should be noted, moreover, that *Gloss-Finder* achieved substantially lower accuracy than ours in identifying the appropriate sign gloss from a submitted video: when the 10 learners were presented with the top **12** most likely matches for their submitted sign (12 being "the maximum number of videos to display in a common monitor resolution without scrolling"), the target sign was found among those 12 only 66% of the time.

For extensive discussion of issues involved in designing this kind of sign search, and factors that affect user satisfaction, see Hassan (2023), which also provides a survey of existing related work.

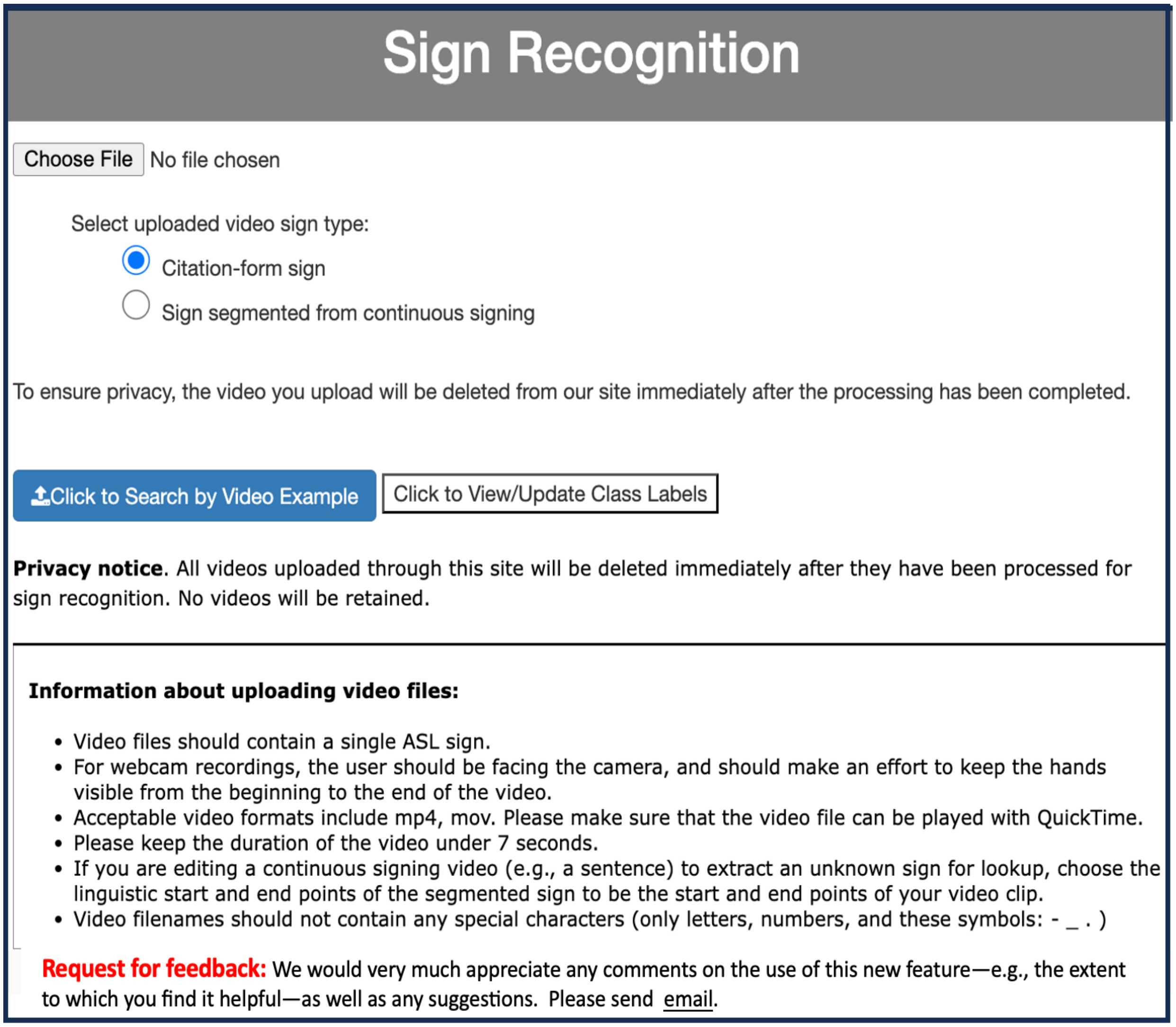

*Figure 2. Interface for Lookup by Video Example*

A couple of illustrations of search results from this system are displayed in Figure 3.

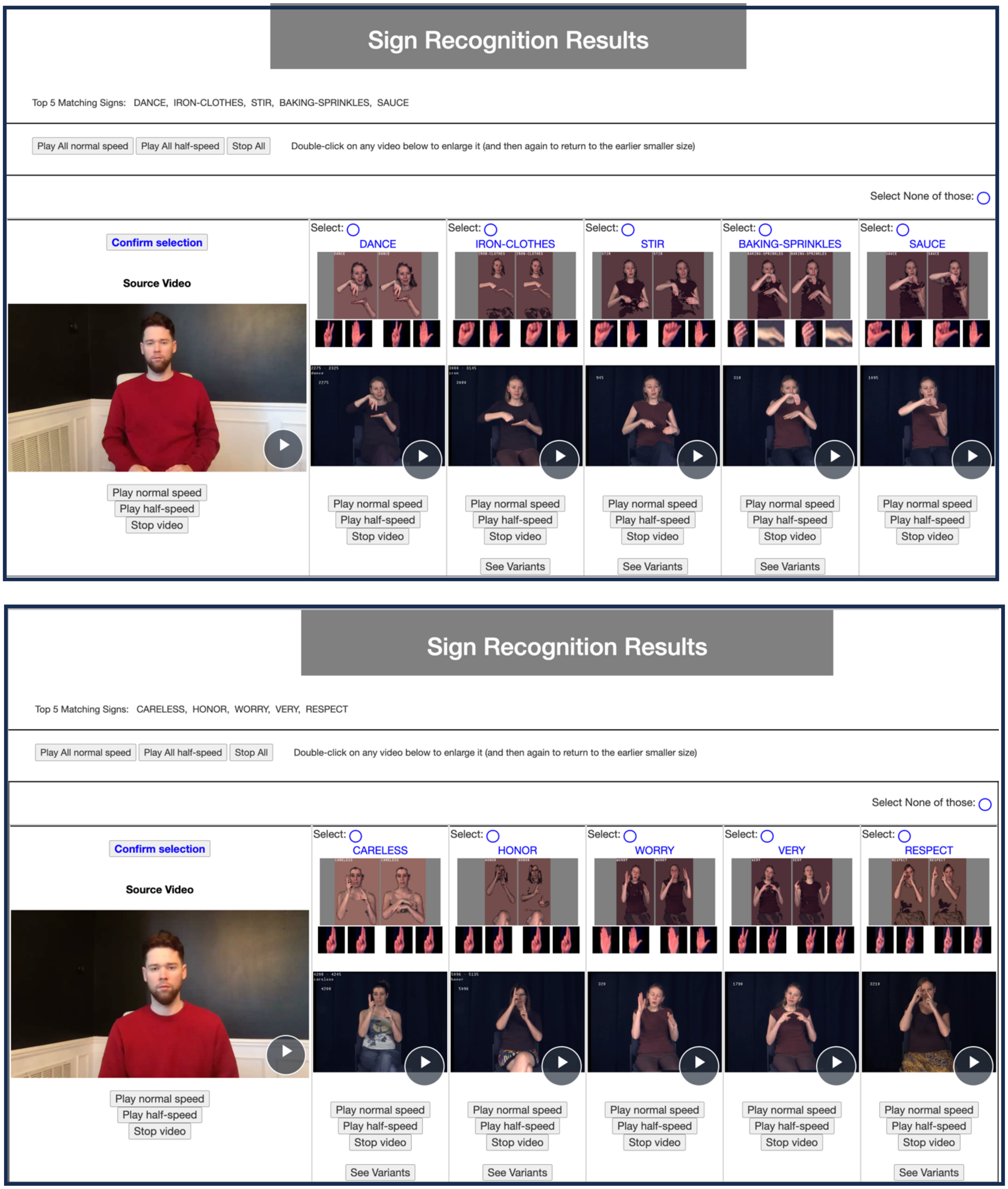

*Figure 3. Sample Results from Sign Lookup from Video Example.*
*In these cases, the correct sign is the leftmost option.*

Here is an example of display of the available variant for AMBULANCE:

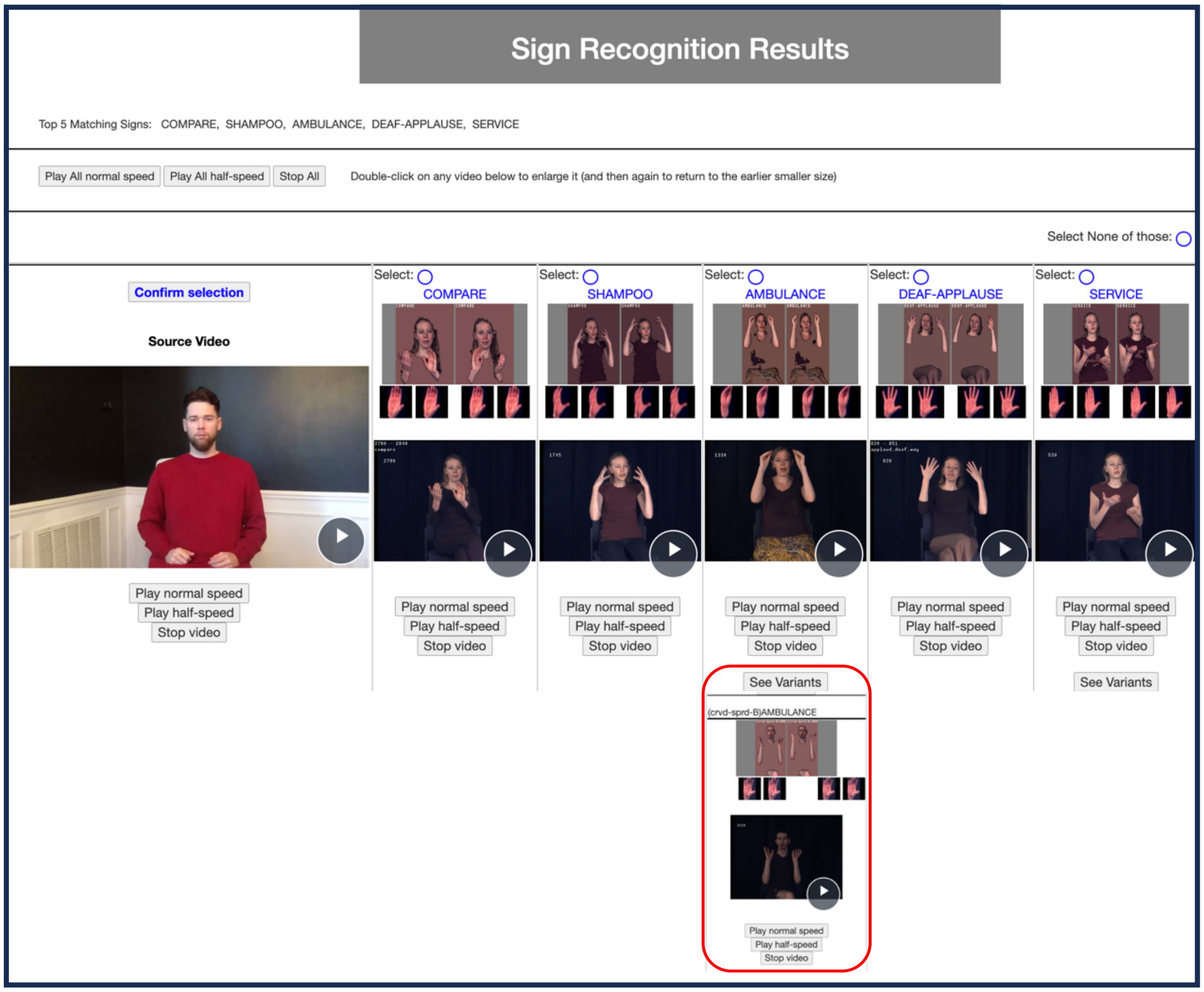

*Figure 4. Sample Display of Sign Variants*

If the selected sign has any variants, the user will be asked to confirm which variant is the right one. This is shown in Figure 5.

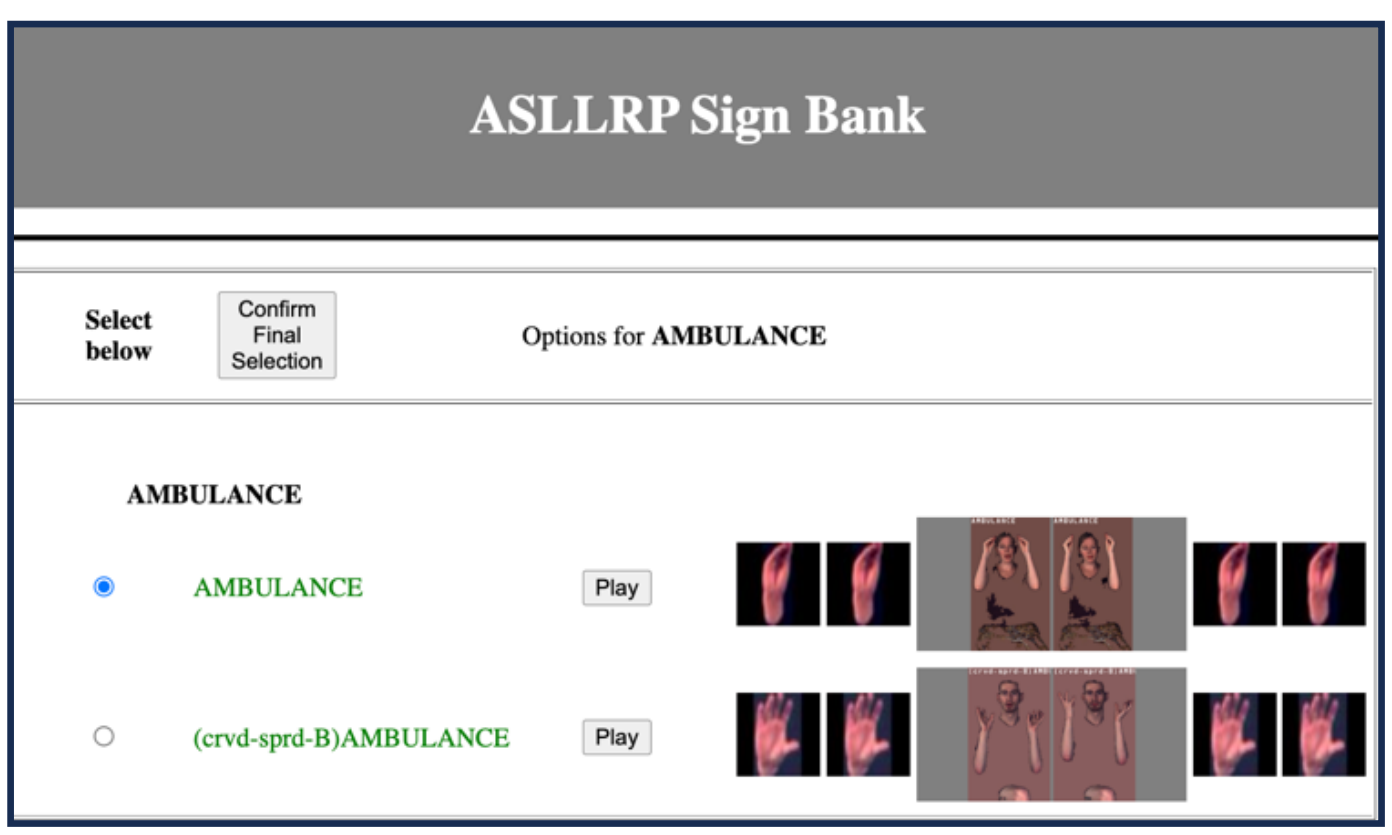

*Figure 5. Display of Results When the Selected Sign has One or More Variants*

Currently, upon confirming the selection, the user will be taken to that entry in our ASLLRP Sign Bank (Neidle, *et al.* 2022b). However, in principle, it would be straightforward to connect this sign recognition module to an ASL dictionary or other resource, so that the user could be taken directly to the appropriate entry. We look forward to collaborating to establish such possible connections in the future.

If the user were to select “COMPARE” from the illustration in Figure 4, the corresponding entry in our Sign Bank (which can be accessed independently) is shown in Figure 6; available video examples, and the utterances from which they came (for signs from sentences), can be played.

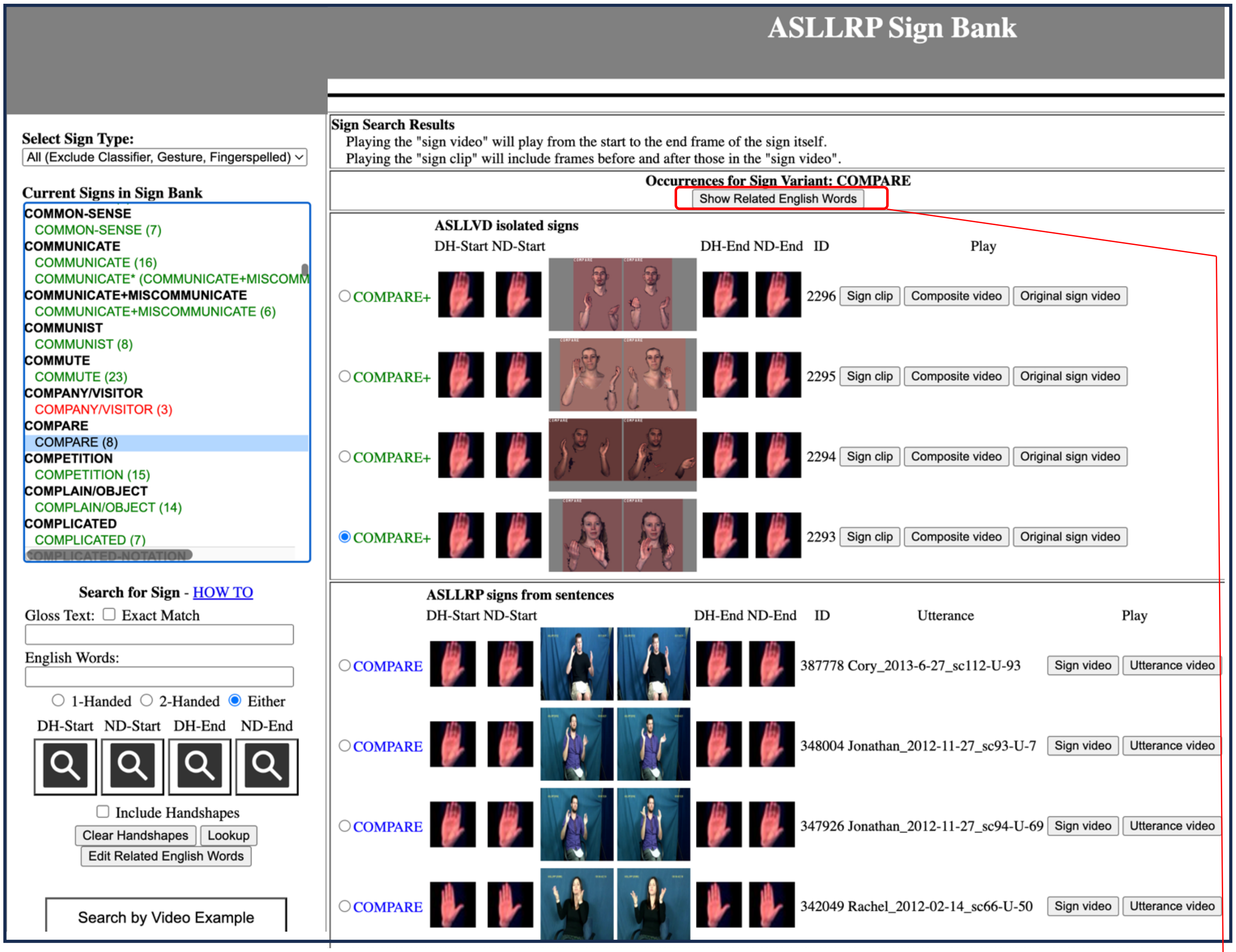

*Figure 6. Sample Entry from our ASLLRP Sign Bank*

Our Sign Bank assigns a unique English-based gloss to each distinct sign production, including lexical variants (Neidle 2002, 2007). This may provide some approximation of meaning; however, it is NOT intended to necessarily reflect an accurate English translation. The user can also choose to “Show Related English Words” (from the top of the entry), as shown in Figure 7 .

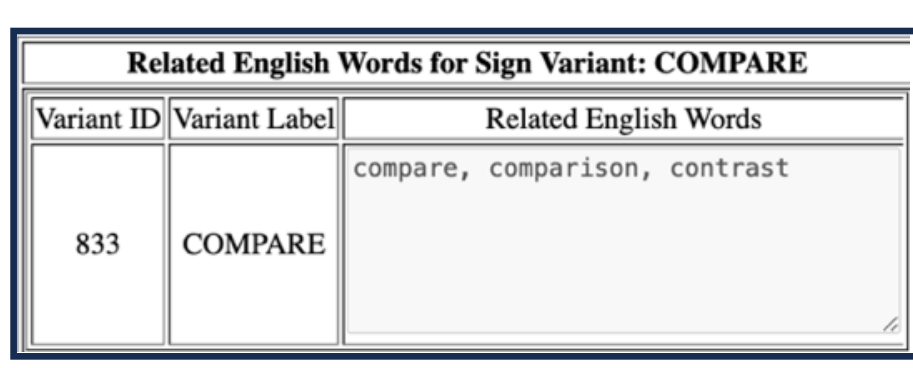

Related English Words for Sign Variant: COMPARE

| Variant ID | Variant Label | Related English Words |
|---|---|---|
| 833 | COMPARE | compare, comparison, contrast |

*Figure 7. Sample Display of Related English Words*

The user may wish to consult other resources for more information about the sign, but the English-based gloss and list of related English words are useful as a starting point for further exploration.

The “search by video example” functionality is available as part of our Sign Bank; these are accessible from <**https://dai.cs.rutgers.edu**>. Note that the ASLLRP Sign Bank can also be searched in other ways, including any combination of: text contained in gloss labels, start/end handshapes, and/or related English words, as shown at the bottom left of Figure 6.

This search function is also integrated into the upcoming release (version 3.5.0) of SignStream®—our software for linguistic annotation of video data (shared from [http://www.bu.edu/asllrp/SignStream/3/](http://www.bu.edu/asllrp/SignStream/3/) free of charge). Thus, a user who is annotating an ASL video can look up an unknown sign directly, and, upon confirmation of the target sign, can then automatically insert the sign’s relevant properties into the SignStream® annotation. This greatly increases efficiency and consistency of annotations.

Versions of SignStream® through 3.4.1 incorporate the ability to search our Sign Bank via gloss text and/or handshape information, and then to insert the selected sign’s properties directly into the annotation (subject to any further editing), as shown in Figure 8.

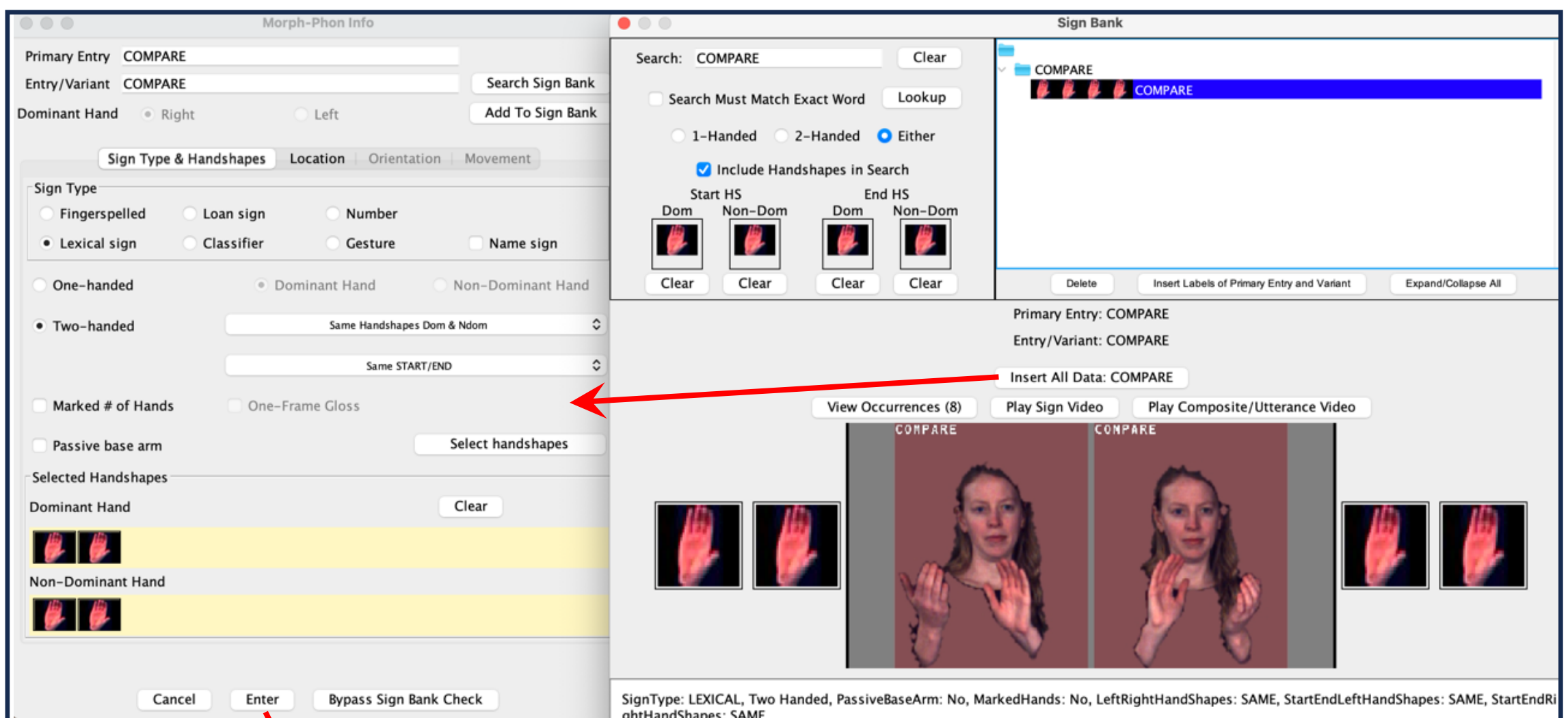

*Figure 8. Sign Bank Search from within SignStream®*

The information can then be directly entered into the main sentence-level annotation, as seen in Figure 9.

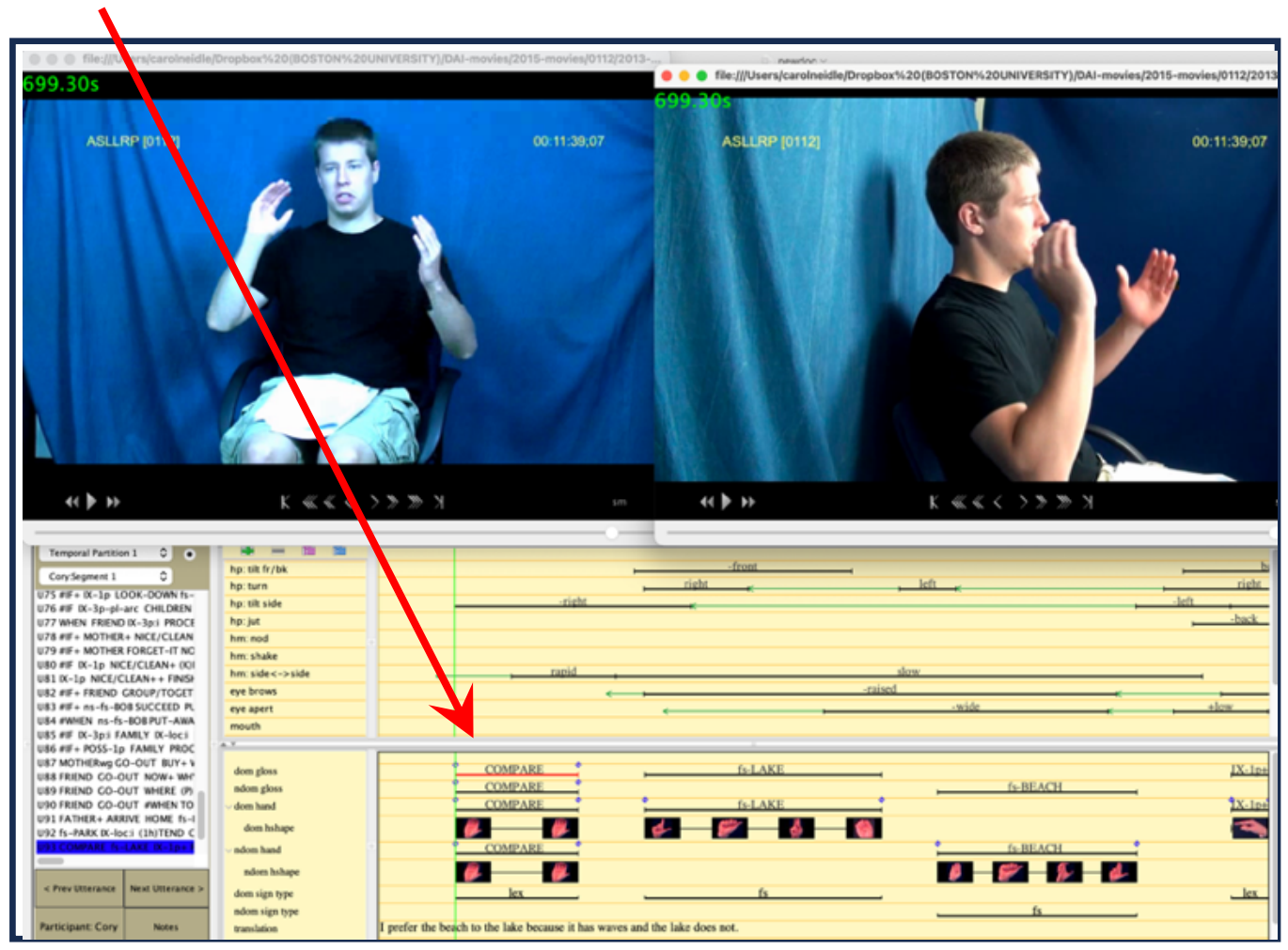

*Figure 9. Sign Information Entered into the Utterance within SignStream®*

The newest version of SignStream® will make it possible, once the user has set start and end points of an unknown sign, to search for that sign through the search-by-video-example module, opening up a window like those shown in Figure 3, and then, upon confirmation of the target sign, to enter the Sign Bank information into the SignStream® annotation.

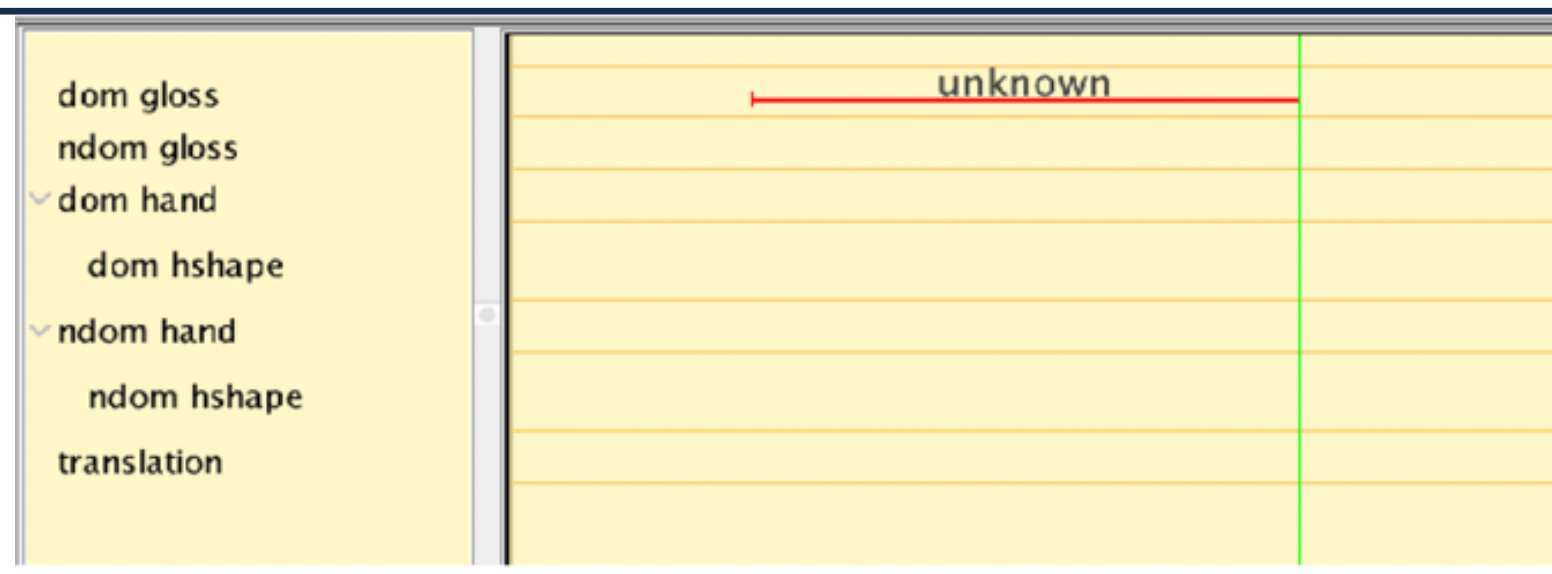

The SignStream user opens the Morph-Phon window and selects "DAI Video Search:

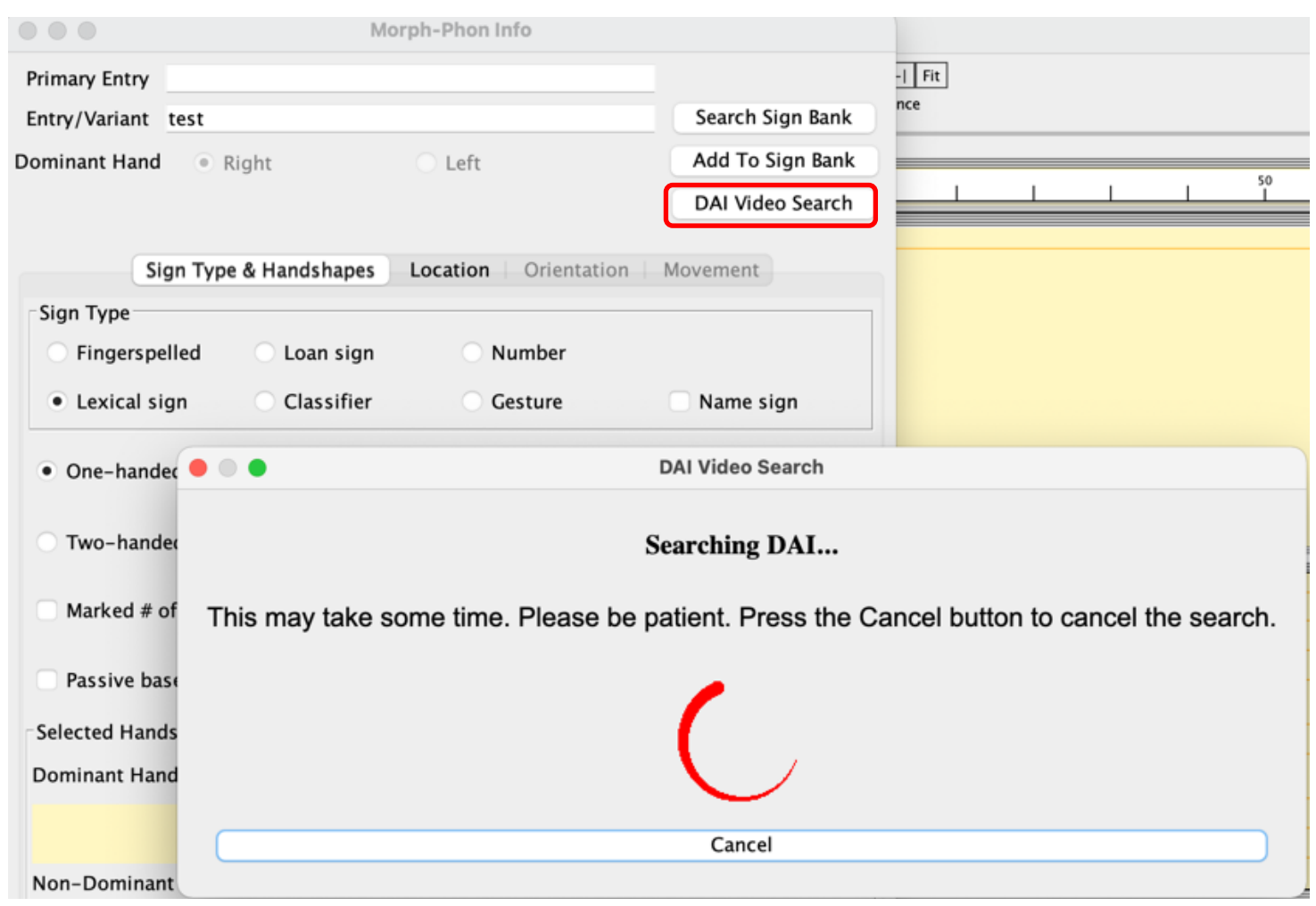

*Figure 10. Brand New Search-by-video-example Feature within SignStream®*

The user will then be presented with the 5 most likely matches for the submitted video, as shown in Figure 3. Upon confirmation of the correct sign, the user can select to "Insert All Data" into that Morph-Phon window, subject to any further editing by the user before the information is entered into the Utterance Window. This is described in greater detail in Neidle and Opoku (2024) and Neidle (2024 (forthcoming)).

### Further Information about the Search-by-video-example Functionality

In all, about 2,360 distinct signs (and similar variants) can currently be recognized via video lookup, including lexical signs, loan signs, numbers, and compounds. We do not currently include fingerspelled signs, classifiers, index signs, and gestures (except when they are included as components of compounds). Our current recognition accuracy for proficient ASL signers is shown in Table 1.

| Type of video input | Top-1 | Top-5 |
|---|---|---|
| Citation-form signs | 81.21 % | 95.36 % |
| Signs segmented from continuous signing | 80.39 % | 92.96 % |

*Table 1. Recognition Accuracy*

Videos of signs from ASL learners, which may differ in production from signs articulated by proficient signers, may be less well recognized. In collaboration with Matt Huenerfauth at RIT, we plan to carry out user studies in the near future to establish the recognition accuracy for ASL learners. We also plan to maintain statistics, as the system is used, for success rates of sign recognition, as previously mentioned.

If the search by video example does not produce the desired result, the user can "Select None of those" (at the upper right in Figure 4), in which case the user will be returned to the main Sign Bank page, and can proceed to search the ASLLRP Sign Bank for the desired sign in other ways: by text contained in the gloss labels, by text in related English words, and/or by handshapes.

For further details, see Neidle and Opoku (2024) and Neidle (2024 (forthcoming)).

### Credits & Acknowledgments

Augustine Opoku has been our Web designer. He is responsible for development and maintenance of our website (https://dai.cs.rutgers.edu/dai/s/dai) through which we share linguistic data and enable the search by video example. The analysis of submitted videos for lookup is carried out behind the scenes by a system that was designed and implemented by Yang Zhou, Xiaoxiao He, and (for our initial version) Konstantinos Dafnis, under the supervision of Dimitris Metaxas. Development of SignStream, our software for linguistic annotation of video data (https://www.bu.edu/asllrp/SignStream/3/), has been carried out principally by Gregory Dimitriadis, at the LCSR (Laboratory for Computer Science Research) at Rutgers University. Carey Ballard has provided assistance and advice on many aspects of these projects, and he has been invaluable in helping with annotations and verifications.

We are grateful to the many people who have helped with the collection, linguistic annotation, and sharing of the ASL data upon which we have relied for this research. In particular, we are endebted to the many ASL signers who have contributed to our database; to Matt Huenerfauth and his team for data collection at RIT; to DawnSignPress for sharing video data; to the many who have helped with linguistic annotations.

This work was supported in part by NSF grants #2235405, #2212302, #2212301, and #2212303, but any opinions, findings, and conclusions expressed in this material are those of the authors and do not necessarily reflect the views of the National Science Foundation.

## Table of Figures